\documentclass[sigconf,nonacm]{acmart}

\AtBeginDocument{%
  }

\acmConference[Conference acronym 'XX]{Make sure to enter the correct
  conference title from your rights confirmation email}{June 03--05,
  2018}{Woodstock, NY}

\acmSubmissionID{291}

\usepackage{booktabs}  
\usepackage{tabularx}  
\usepackage{array}
\usepackage{amsmath}
\newcolumntype{Y}{>{\centering\arraybackslash}X}
\usepackage{colortbl}  
\usepackage{bm}
\usepackage{graphicx}
\usepackage{subcaption}
\usepackage{float}
\usepackage[dvipsnames]{xcolor}
\usepackage[table]{xcolor}
\definecolor{bg_blue}{RGB}{228, 240, 255}
\definecolor{rose_red}{RGB}{219, 48, 122} 
\definecolor{bg_gray}{RGB}{240, 240, 240}
\definecolor{bg_pink}{RGB}{255, 240, 245} 
\definecolor{dark_bg_blue}{RGB}{153, 204, 255} 
\definecolor{bg1_blue}{RGB}{204, 229, 255}
\definecolor{cite_blue}{RGB}{70, 105, 170}

\begin{document}

\title{GAMR: Geometric-Aware Manifold Regularization with Virtual Outlier Synthesis for Learning with Noisy Labels}

\author{Ningkang Peng}
\affiliation{%
  \institution{Nanjing Normal University}
  \city{Nanjing}
  \country{China}}
\author{Jingyang Mao}
\affiliation{%
  \institution{Nanjing Normal University}
  \city{Nanjing}
  \country{China}}
\author{Xiaoqian Peng}
\affiliation{%
  \institution{Nanjing University of Chinese Medicine}
  \city{Nanjing}
  \country{China}}
\author{Peirong Ma}
\affiliation{%
  \institution{Nanjing Normal University}
  \city{Nanjing}
  \country{China}}
\author{Xichen Yang}
\affiliation{%
  \institution{Nanjing Normal University}
  \city{Nanjing}
  \country{China}}
\author{Weiguang Qu}
\affiliation{%
  \institution{Nanjing Normal University}
  \city{Nanjing}
  \country{China}}
\author{Yanhui Gu}
\affiliation{%
  \institution{Nanjing Normal University}
  \city{Nanjing}
  \country{China}}








\renewcommand{\shortauthors}{Trovato et al.}

\begin{abstract}
Deep neural networks (DNNs) experience significant performance degradation when processing noisy labels, primarily due to overfitting on mislabeled data. Current mainstream approaches attempt to mitigate this issue by passively filtering clean samples during training. However, simple sample filtering within feature spaces degraded by noise struggles to distinguish between challenging samples and noisy samples, creating a bottleneck for model performance. We highlight for the first time the fundamental importance of actively reshaping feature space geometry for learning from noisy data. We propose a novel Geometry-aware Manifold Regularization Paradigm whose core idea is to explicitly construct energy barriers between data manifolds by actively synthesizing virtual outlier samples. By imposing geometric constraints that promote intra-class compactness and inter-class separation, this approach enhances the discriminability between hard and noisy samples, leading to the learning of more robust representations. Our regularization mechanism exhibits high universality, with effectiveness independent of any prior assumptions about noise patterns. It can be integrated as a standalone mechanism into existing sample selection frameworks, providing stronger robustness against diverse noisy environments. Experiments demonstrate that our paradigm achieves performance surpassing current state-of-the-art (SOTA) methods on multiple benchmarks, including CIFAR-10, with particularly pronounced advantages under more challenging asymmetric noise conditions. Furthermore, this paradigm significantly enhances the model's capability in Out-of-Distribution (OOD) detection, ensuring superior reliability and safety for deployment in open-world scenarios.
\end{abstract}

\maketitle


\section{Introduction}
In recent years, DNNs have shown impressive performance in computer vision tasks\cite{medical2}. However, large-scale,high-quality labeled data is necessary for these achievements, while label noise is practically inevitable in practical applications. Models frequently overfit to wrong labels when trained with noisy labels, seriously impairing generalization performance\cite{performence3}.

Two major categories can be used to classify current methods for Learning with Noisy Labels (LNL). One class of methods mitigates noise effects by estimating noise transfer matrices or performing statistical corrections\cite{matrix4,li2025learning}. Nevertheless, these approaches typically rely on robust distributional assumptions and perform poorly under high noise rates or scenarios with a large number of categories\cite{liao2025instance}. In contrast, the sample selection-based paradigm has emerged as the mainstream approach\cite{zhang2025learning,kim2025splitnet}. The majority of these techniques dynamically divide data into clean and noisy subsets by fitting the distribution of training losses\cite{unsupervised}, then use semi-supervised learning (SSL) to optimize these divisions\cite{mixmatch,manifoldDivideMix}.

Although sample selection-based methods have demonstrated their effectiveness in practice, relying solely on loss or prediction confidence as selection criteria fails to fundamentally distinguish between closely related clean samples and noisy samples. This limitation caps the upper bound of sample selection accuracy and ultimately constrains model generalization.

\begin{figure}[t]  
    \centering
    \includegraphics[width=1\linewidth]{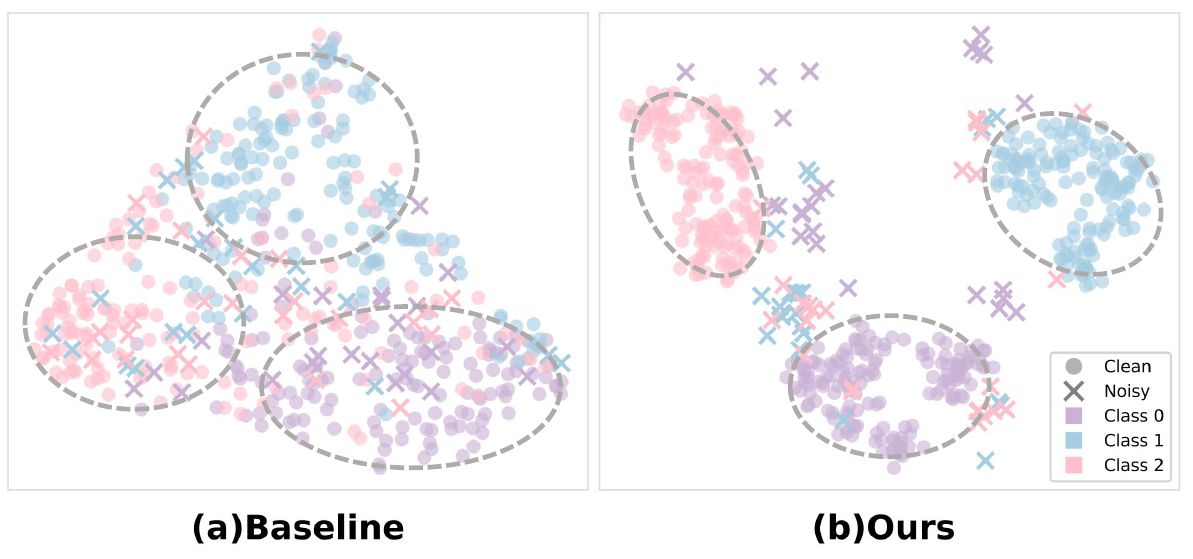} 
    \caption{\textbf{Feature Visualization Comparison.} (a) Exhibits a loose structure with entangled clean and noisy samples. (b) Shows compact class distributions with effective noise segregation.}
    \label{fig:tu}
\end{figure}

For the first time, this paper shifts the research perspective to actively reshaping the feature space for sample selection, and proposes a Geometry-aware Manifold Regularization Paradigm. Unlike previous methods that directly distinguish samples in a loose feature space, our goal is to actively construct a feature space that is compact within classes and separable between classes\cite{supervised}. Specifically, we dynamically synthesize virtual outlier samples in low-density regions of the feature space and impose explicit high-energy constraints through energy modeling\cite{VOS}. This establishes distinct energy barriers between low-energy class clusters\cite{energy}. By imposing anisotropic constraints on the real data manifold, we force samples within the same class to cluster tightly around their class centers while separating different classes. As shown in Figure~\ref{fig:tu}, this structural reshaping of the feature space ultimately enables effective distinction between difficult samples and noisy samples that originally overlapped in distribution. This enhances the diversity of sample selection, allowing the model to learn a more comprehensive and diverse set of difficulty clean samples. Crucially, this geometric correction does not rely on explicit modeling of noise patterns but functions as a universal regularization mechanism that can be integrated into existing sample selection frameworks. Our contributions are summarized as follows:
\begin{itemize}
\item \textbf{We propose a novel geometric paradigm for LNL.} From a geometric perspective, we reveal that the performance bottleneck of the current mainstream passive selection paradigm stems from operating within a feature space degenerated by noise. This paper pioneers a paradigm shift, redirecting research on learning with noise toward active feature space reshaping.

\item \textbf{We design an energy-based universal manifold compactification mechanism.} We propose a noise pattern-agnostic regularization mechanism that enhances the noise robustness of sample selection methods by synthesizing high-energy virtual outlier samples in low-density regions of the feature space.


\item \textbf{Enhancing convergence efficiency, performance, and trustworthy robustness of noisy training.} Extensive experiments substantiate that this paradigm significantly enhances the separability of clean and noisy samples in the feature space, achieving state-of-the-art (SOTA) results on multiple benchmarks, including CIFAR-10/100, especially under challenging asymmetric noise conditions. Furthermore, it guides the model to converge more rapidly onto the clean data manifold, accelerating the overall training process. Crucially, distinguishing our work from prior arts limited to in-distribution (ID) accuracy, we extend the evaluation scope of learning with noisy labels. We reveal, for the first time, that the active geometric reshaping paradigm exhibits exceptional safety against unknown samples, ensuring high generalization capability in open-world scenarios.
\end{itemize}

\section{Related Work}
\subsection{Labeled Noise Learning}

Existing sample selection mechanisms in LNL can be broadly categorized into small-loss criterion-based methods and hybrid SSL frameworks\cite{cheng2025exploring}. Early approaches leveraged the empirical observation that DNNs prioritize learning clean patterns, thereby selectively retaining reliable samples\cite{Gui13}. For instance, Han et al\cite{Han9}. proposed a Co-teaching architecture employing two structurally identical DNNs to mutually filter clean samples based on Cross-Entropy (CE) loss\cite{CE}. Recent methodologies, such as DivideMix\cite {li2020dividemix}, MOIT\cite {MOIT14}, and DISC\cite {DISC18}, have evolved beyond merely discarding residual samples, instead repurposing them as unlabeled data. By integrating strategies like Gaussian Mixture Models, contrastive loss, and dynamic thresholds, these methods optimize training through SSL, achieving significant performance gains. Building on these frameworks, subsequent studies have incorporated more diverse theoretical tools to enhance robustness. For example, Feng et al.\cite{Feng19} refined sample selection from the perspective of optimal transport theory, integrating it with the DivideMix pipeline. 
Although these approaches achieved some success in practical applications, their samples were passively selected within unconstrained, chaotic feature spaces. Consequently, the upper bound of sample selection accuracy is bottlenecked, which ultimately constrains the model's generalization performance.

\subsection{Outlier Exposure}

To delineate clear decision boundaries within the feature space, Hendrycks et al. introduced the concept of Outlier Exposure (OE)\cite{OE}, which actively incorporates non-target data during training. However, this approach relies on auxiliary OOD datasets that are often challenging to acquire. Advancing this further, Virtual Outlier Synthesis (VOS)\cite{VOS} employs a probabilistic modeling strategy by estimating class-conditional Gaussian distributions in the feature space and sampling virtual outliers from them. While these methodologies have primarily been employed to detect unknown external data, our work marks the first application of this mechanism to address the internal data pollution problem inherent in Noisy Label Learning.
\begin{figure*}[t]
    \centering
    \includegraphics[width=0.99\textwidth]{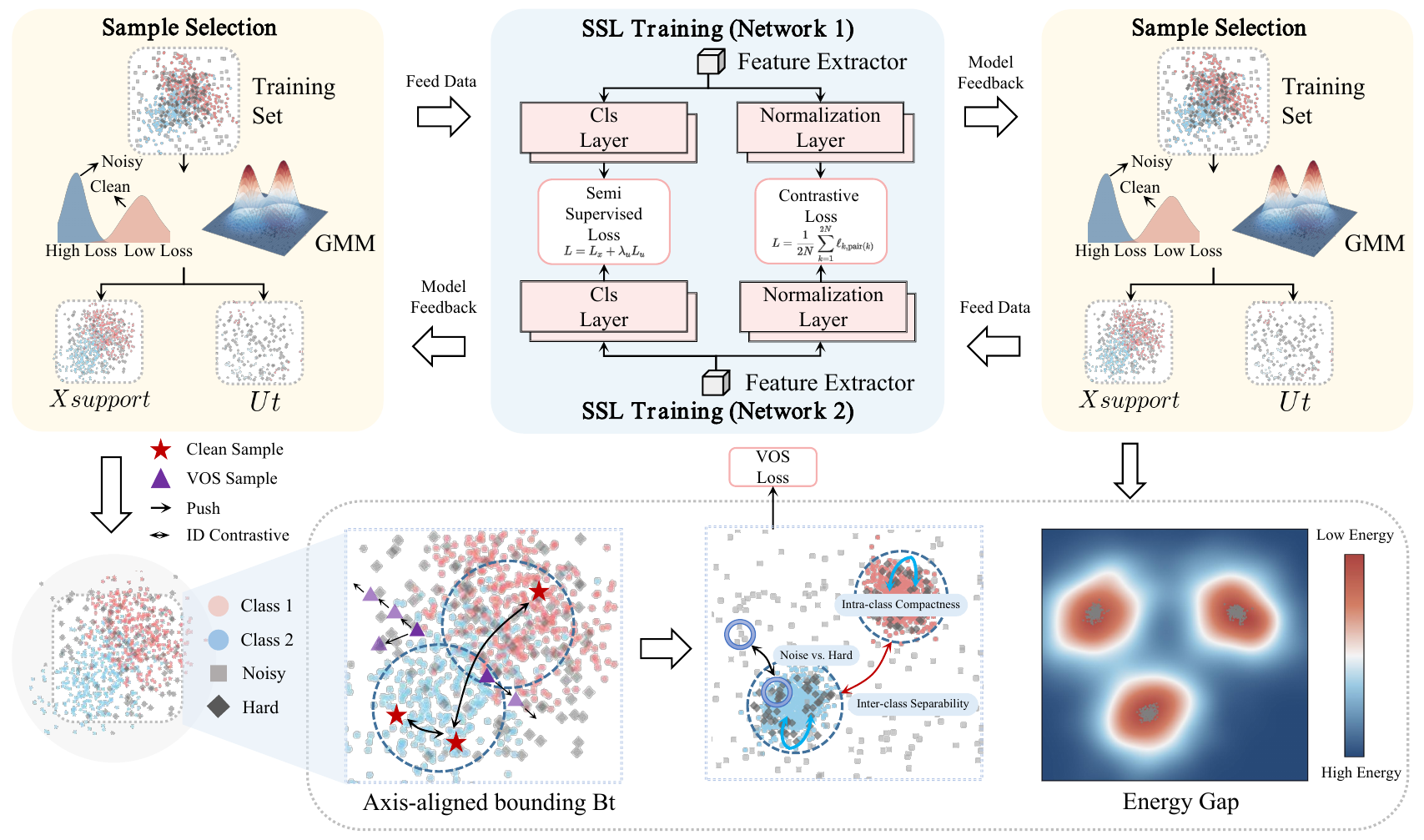} 
    \caption{\textbf{The overall architecture of the Two-Stage Repair Framework.} The framework comprises two synergistic components: \textbf{(Top) Sample Selection and SSL Training:} Dual networks collaborate via GMM to partition data into a clean set $\mathcal{X}_{\text{support}}$ and an unlabeled set $\mathcal{U}_t$, followed by semi-supervised contrastive training. \textbf{(Bottom) Geometry-Aware Manifold Regularization:} Based on $\mathcal{X}_{\text{support}}$, dynamic axis-aligned bounding boxes $\mathcal{B}_t$ are constructed to synthesize virtual outliers.}
    \label{fig:1}
\end{figure*}
\section{Method}

Aligning with the prevailing partition-then-train paradigm\cite{zhang2025psscl}, our framework operates in two progressive stages, initially distilling a high-confidence clean subset from the noisy data, and subsequently expanding this set to facilitate robust training for improved model performance.
\subsection{Warm-up and Geometric Screening}

\subsubsection{Dual Network Warm-up}

We initialize two DNNs with identical structures, denoted as $f_1$ and $f_2$, each consisting of a feature extractor $h(\cdot; \phi)$ and a classifier $f(\cdot; \theta)$. The dual networks mitigate the judgment bias inherent in a single model through cross-validation and mutual screening mechanisms\cite{Han9}. All training and selection operations are performed on the noisy training set $\tilde{D} = \{(x_i, \tilde{y}_i)\}_{i=1}^{n}$. The overall architecture of our proposed framework is depicted in Figure~\ref{fig:1}.
To enhance the model's universality and robustness against various types of label noise, and to ensure the early stability of the manifold structure, we adopt the Generalized Cross Entropy (GCE) loss\cite{GCE} during the warm-up phase.

\subsubsection{Geometric Support Set Collection}

High-confidence samples filtered based on latent features precisely characterize intra-class distributions, serving as a robust basis for virtual sample synthesis. This collection process is implemented via a two-step strategy comprising coarse data partitioning using a two-component GMM and high-confidence purification. The specific procedure is detailed as follows:

First, we model the GCE loss values for all training samples. Following the small-loss criterion, we compute the clean probability $w_i$ for each input sample $x_i$.
\begin{equation}
w_{i}^{t} = p(g \mid l_{GCE}),
\end{equation}
where $g$ corresponds to the component with the smaller mean in the GMM, and $l_{GCE}$ represents the GCE loss value for each training sample. Based on this clean probability, we utilize the following formulas to partition the entire training set into a high-confidence labeled set and a low-confidence unlabeled set, thereby completing the coarse screening of the manifold support set. The partition formulas are defined as follows:
\begin{equation}
\begin{aligned}
\mathcal{X}_t &= \left\{ (x_i, \tilde{y}_i, w_i^t) \mid w_i^t \ge \tau, \ \forall (x_i, \tilde{y}_i) \in \tilde{D} \right\},
\end{aligned}
\end{equation}
\begin{equation}
\begin{aligned}
\mathcal{U}_t &= \left\{ (x_i, w_i^t) \mid w_i^t < \tau, \ \forall (x_i, \tilde{y}_i) \in \tilde{D} \right\},
\end{aligned}
\end{equation}
where $t$ denotes the current training epoch, $\mathcal{X}_t$ represents the labeled set with high label credibility, and $\mathcal{U}_t$ represents the unlabeled set.
However, considering that sample loss values may fluctuate in high-noise scenarios, a single GMM partition is insufficient to ensure the purity of the candidate set. To address this issue, we employ a long-term confidence detection mechanism, selecting only those samples that have been judged as labeled samples based on the aforementioned formulas for $v$ consecutive training epochs.
{\small
\begin{align}
\mathcal{X}_t &= \left\{ (x_i, \tilde{y}_i, w_i^t) \mid \sum_{e=t-v+1}^{t} \mathbf{1}(w_i^e \ge \tau) = v, \ \forall (x_i, \tilde{y}_i) \in \tilde{D} \right\},
\end{align}
}
{\small
\begin{equation}
\begin{aligned}
\mathcal{U}_t &= \left\{ (x_i, w_i^t) \mid \sum_{e=t-v+1}^{t} \mathbf{1}(w_i^e \ge \tau) \neq v, \ \forall (x_i, \tilde{y}_i) \in \tilde{D} \right\}.
\end{aligned}
\end{equation}
}

Here, $\mathbf{1}(\cdot)$ is the indicator function. By adopting this sliding window mechanism, we ensure the reliability of sample labels while eliminating outlier samples whose features deviate from the manifold, ultimately forming the high-confidence manifold support set $\mathcal{X}_{\text{support}}$.

\subsection{Geometry-aware Manifold Regularization}

\subsubsection{EVT-Guided Manifold Support Estimation}

Sampling virtual outliers directly from the entire unbounded feature space is both computationally intractable and geometrically ill-defined. Instead, our objective is to estimate the finite support $\mathcal{X}_{\text{support}}$ of the underlying data manifold. Grounded in Extreme Value Theory (EVT), specifically the Fisher-Tippett-Gnedenko theorem, we leverage the fundamental property that the distribution of sample extremes asymptotically converges to the Generalized Extreme Value (GEV) distribution $G(z)$, independent of the parent data distribution.
\begin{equation}
    \lim_{n \to \infty} P\left(\frac{M_n - \mu_n}{\sigma_n} \le z\right) = G(z) = \exp\left( -\left( 1 + \xi z \right)^{-1/\xi} \right),
\end{equation}
where $M_n$ denotes the sample maximum, $\mu_n$ and $\sigma_n$ are normalizing constants, and $\xi$ is the shape parameter. In the context of bounded feature representations, the distribution falls into the \textit{Weibull} domain of attraction ($\xi < 0$), which implies the existence of a finite upper endpoint $\omega < \infty$. This provides a rigorous statistical guarantee that the boundaries of the latent manifold can be effectively characterized by the tail extrema of feature projections.
For each training epoch $t$, we map all $N$ samples from $\mathcal{X}_{\text{support}}^{(t)}$ into the $d$-dimensional feature space, thereby obtaining a global feature set $\mathcal{Z}_{\text{clean}}^{(t)}$, where each feature vector $\mathbf{z}_i \in \mathbb{R}^d$, this set represents the core distribution of the manifold at the current stage:
\begin{equation}
    \mathcal{Z}_{\text{clean}}^{(t)} = \{\mathbf{z}_1, \mathbf{z}_2, \dots, \mathbf{z}_N\}.
\end{equation}
To capture the geometric span of the class distribution without assuming a parametric density, we decompose the $d$-dimensional space into independent marginal projections. For an arbitrary feature dimension $j$ (where $j \in \{1, \dots, d\}$), we first extract the coordinate values of all $N$ samples along this dimension to construct a scalar set $C_j$:
\begin{equation}
    C_j = \{z_{1,j}, z_{2,j}, \dots, z_{N,j}\}.
\end{equation}

Here, $z_{i,j}$ denotes the $j$-th component of the feature vector $\mathbf{z}_i$. The set $C_j$ characterizes the distribution of all clean samples along the $j$-th feature dimension.

To comprehensively estimate the feature distribution range, we model the boundaries of the latent manifold at each epoch using EVT. By extracting the component-wise sample extrema $b_{\min,j}^{(t)} = \min(C_j)$ and $b_{\max,j}^{(t)} = \max(C_j)$ from the feature dimension set $C_j$, capture the geometric limits of the class expansion in the feature space.
Leveraging the asymptotic guarantees provided by EVT, we construct a $d$-dimensional hyper-rectangular envelope $\mathcal{B}_t$. This region represents the maximal possible connected domain of the samples within $\mathcal{X}_{\text{support}}$ in the feature space. Unlike traditional Gaussian Mixture Models (GMM), this modeling approach possesses intrinsic immunity to label noise residing in the high-density interior, as it relies exclusively on the tail properties of the distribution.
\begin{equation}
    \mathcal{B}_t = \left\{ \mathbf{z} \in \mathbb{R}^d \mid \mathbf{b}_{\min}^{(t)} \preceq \mathbf{z} \preceq \mathbf{b}_{\max}^{(t)} \right\}.
\end{equation}

All subsequent outlier candidate sets $\mathcal{Z}_{\text{cand}}$ are generated via uniform sampling within this bounded region $\mathcal{B}_t$.

\subsubsection{Geometric Outlier Filtering}

The EVT-derived envelope $\mathcal{B}_t$ delineates the outer limits of the feature distribution. However, the interior of $\mathcal{B}_t$ contains both the high-density ID manifold and the low-density uncertainty regions. To specifically target the latter for virtual outlier synthesis, we employ a geometric filtering strategy that acts as a complement to the EVT boundary. To mathematically implement this criterion, we define the geometric center $\mu_c$ as a first-order approximation of the class position within the feature space $\mathbb{R}^d$.
\begin{equation}
\mu_{c}^{(t)} = \frac{1}{|\mathcal{Z}_{c}^{(t)}|} \sum_{z \in \mathcal{Z}_{c}^{(t)}} z.
\end{equation}

In the $t$-th training epoch, the geometric centers of all categories $\{\mu_1^t, \mu_2^t, \dots, \mu_K^t\}$ collectively constitute the center set $P_t$. For any arbitrary point $z$ within the candidate set $\mathcal{Z}_{\text{cand}}$, we compute its Euclidean distances to all geometric centers in $P_t$ and determine the minimum value. This minimum distance, denoted as $d_{\min}(z, P_t)$, serves to quantify the proximity of the candidate point to its nearest class core:
\begin{equation}
    d_{\min}(z, P_t) = \min_{c \in \{1, \dots, K\}} \|z - \mu_c^t\|_2.
\end{equation}

A candidate point $z$ is accepted as a valid virtual outlier and included in the final set $\mathcal{V}_{\text{vos}}$ if and only if its minimum distance exceeds a predefined rejection radius $\tau$. Consequently, the final set of synthesized outliers is formally defined as:
\begin{equation}
    \mathcal{V}_{\text{vos}} = \{z \in \mathcal{Z}_{\text{cand}} \mid \min_{c \in \{1, \dots, K\}} \|z - \mu_c^t\|_2 > \tau \}.
\end{equation}
Geometrically, this construction strategy targets the hollow space between the class centroid and the EVT-derived boundary. These synthesized samples effectively simulate the uncertainty boundary, compelling the model to delineate a compact decision surface that is strictly confined between the high-confidence manifold core and the EVT limits.

\subsubsection{Energy-Based Boundary Solidification}

The geometric rejection criterion provides a discrete set of virtual synthetic samples $\mathcal{V}_{\text{vos}}$. A key challenge lies in transforming this discrete geometric information into a dynamic and differentiable constraint, thereby actively forming energy barriers during model training to distinguish between hard samples and noisy samples. To address this issue, we introduce the perspective of Energy-Based Models (EBMs)\cite{EBM}. EBMs map each point $\mathbf{z}$ in the feature space to a scalar energy value. Specifically, let $f(\mathbf{z}; \theta)$ be the classifier head mapping the feature $\mathbf{z}$ to $K$ class logits. The energy function $E(\mathbf{z}; \theta)$ is defined as:
\begin{equation}
    E(\mathbf{z}; \theta) = -T \cdot \log \sum_{k=1}^K \exp\left(\frac{f_k(\mathbf{z}; \theta)}{T}\right),
\end{equation}
where $T$ is the temperature parameter. Based on the Boltzmann distribution $p(\mathbf{z}) \propto \exp(-E(\mathbf{z}; \theta))$, a lower energy value indicates a higher probability density. Consequently, the objective of model training transforms into shaping the manifold region containing clean samples ($\mathcal{Z}_{\text{clean}}$) into a low-energy potential well, while simultaneously shaping the regions occupied by the identified virtual outliers ($\mathcal{V}_{\text{vos}}$) between classes into high-energy barriers.
\begin{equation}
    p(z;\theta) = \frac{e^{-E(z;\theta)/T}}{Z(\theta)}.
\end{equation}

Here, $ Z(\theta) = \int_{z} e^{-E(z;\theta)/T} dz$ serves as the partition function to ensure the probability distribution is normalized.

In this framework, the regularization loss $\mathcal{L}_{\text{spade}}$ requires the model not only to perform classification but also to discriminate whether a sample originates from the high-density manifold core or a low-density region. This objective is realized through a hyperparameter-free Binary Cross Entropy (BCE) loss:
\begin{equation}
\begin{split}
    \mathcal{L}_{\text{spade}} = \, & \mathbb{E}_{x \sim \mathcal{Z}_{\text{clean}}} [-\log(1-\sigma(E(x)))] \\
    & + \mathbb{E}_{v \sim \mathcal{V}_{\text{vos}}} [-\log(\sigma(E(v)))],
\end{split}
\end{equation}
where $\sigma(\cdot)$ denotes the Sigmoid function, which maps the unbounded energy value $E(z)$ to a probability estimate that $z$ belongs to the out-of-distribution domain. The first term forces the model to minimize the energy $E(x)$ of clean samples by maximizing their probability. Conversely, the second term drives the maximization of the energy $E(v)$ for virtual outliers by maximizing their probability.

This energy-based synergistic effect creates an anisotropic gradient field. It forces all samples belonging to the same category to aggregate highly towards their class core, rendering the entire class cluster exceptionally compact. Geometrically, this widens the distance between hard samples and noisy samples, achieving separability between them. This enables the model to learn a more comprehensive and diverse set of clean samples, rather than being limited to simple samples.

\subsection{Semi-Supervised Contrastive Learning}
\subsubsection{Dual-Objective Optimization} 

We perform collaborative label optimization and data augmentation on the candidate labeled set $\mathcal{X}_{\text{support}}$ and the unlabeled set $\mathcal{U}_t$. Specifically, for samples in $\mathcal{X}_{\text{support}}$, we rectify the labels by combining clean probabilities with model predictions; for samples in $\mathcal{U}_t$, we generate pseudo-labels via a sharpening operation on the predictions from the dual networks.
Subsequently, we compute the standard SSL loss. This loss comprises the labeled set loss $l_x$, the unlabeled set loss $l_u$, and a regularization term $l_{reg}$, which are modulated by the balancing coefficients $\lambda_u$ and $\lambda_{reg}$:
\begin{equation}
    \mathcal{L}_{\text{ssl}} = \frac{1}{|\mathcal{X}_{t}^{\prime}|}\sum l_{x} + \frac{\lambda_{u}}{|\mathcal{U}_{t}^{\prime}|}\sum l_{u} + \lambda_{reg}l_{reg}.
\end{equation}

To learn more discriminative representations from the unlabeled data, we impose an unsupervised contrastive loss $\mathcal{L}_{\text{cl}}$ on the unlabeled set $\mathcal{U}_t$. By pulling together representations of differently augmented views of the same image while pushing apart representations of distinct images, this loss mitigates the dependence of the model on potentially erroneous pseudo-labels:
\begin{equation}
    l_{\text{cl}} = -\frac{1}{2|\mathcal{U}_{t}^{\prime}|} \cdot \sum_{i=1}^{\frac{1}{2}|\mathcal{U}_{t}|} [l(2i-1, 2i) + l(2i, 2i-1)].
\end{equation}

The similarity score function $l(n, m)$ is defined as follows:
\begin{equation}
    l(n, m) = \log \frac{\exp(sim(z_{n}, z_{m})/\delta)}{\sum_{k=1, k \ne n}^{2N_{b}} \exp(sim(z_{n}, z_{k})/\delta)}.
\end{equation}

Here, $z_n$ and $z_m$ denote the normalized representations in a low-dimensional space. These are obtained by passing two views, $x_n$ and $x_m$—derived from each sample $x_i \in \mathcal{U}_t$ via two independent strong data augmentations—through the feature extractor $h(\cdot)$ and an additional projection head $g(\cdot)$.

\subsubsection{Total Loss Function}

Finally, by incorporating the regularization loss $\mathcal{L}_{\text{spade}}$ introduced in Section Dual-Objective Optimization, the total loss function $\mathcal{L}_{\text{total}}$ integrates standard SSL, unsupervised representation learning, and our proposed active geometric correction into a unified optimization objective at each training step:
\begin{equation}
    \mathcal{L}_{\text{total}} = \mathcal{L}_{\text{ssl}} + \lambda_{\text{cl}} l_{\text{cl}} + \lambda_{\text{spade}} \mathcal{L}_{\text{spade}},
\end{equation}
where $\lambda_{\text{cl}}$ and $\lambda_{\text{spade}}$ serve as hyperparameters to balance the contribution of different regularization terms. By jointly optimizing this total loss, our framework effectively distinguishes between noisy samples and hard samples, thereby enhancing the diversity of sample selection. This enables the model to learn a more comprehensive and diverse set of hard clean samples, ultimately achieving highly robust learning against noisy data.

\begin{table*}[t]
    \caption{Experimental results on CIFAR-10 dataset.}
    \label{tab:cifar10_results}
    
    \centering
    \begin{tabularx}{\textwidth}{l *{7}{Y}}
        \toprule
        Dataset & \multicolumn{6}{c}{CIFAR-10} & \\ 
        Noise Mode & \multicolumn{4}{c}{Sym.} & \multicolumn{2}{c}{Asym.} & \\ 
        
        Method & 20\% & 50\% & 80\% & 90\% & 40\% & 49\% & Avg. \\ 
        \midrule
        
        \rowcolor{bg_gray}
        Standard CE & 86.8 & 79.4 & 62.9 & 42.7 & 85.0 & - & 71.4 \\
        
        \rowcolor{bg_gray}
        GCE & 86.6 & 81.9 & 54.6 & 21.2 & 76.0 & - & 64.1 \\
        
        \rowcolor{bg_gray}
        GCE+CL & 90.0 & 89.3 & 73.9 & 36.5 & 78.1 & - & 73.6 \\
        
        \rowcolor{bg_blue}
        DivideMix & 96.1 & 94.6 & 93.2 & 76.0 & 93.4 & 83.7 & 90.7 \\
        
        ELR+ & 95.8 & 94.8 & 93.3 & 78.7 & 93.0 & - & 91.1 \\
        MOIT+ & 94.1 & 91.8 & 81.1 & 74.7 & 93.3 & - & 87.0 \\
        Sel-CL+ & 95.5 & 93.9 & 89.2 & 81.9 & 93.4 & - & 90.8 \\
        
        \rowcolor{bg_gray}
        MixUp & 95.6 & 87.1 & 71.6 & 52.2 & 77.7 & - & 76.8 \\
        UNICON & 95.1 & 93.7 & 92.1 & 90.8 & 91.7 & - & 92.7 \\
        
        SOP+ & 96.3 & 95.5 & 94.0 & - & 93.8 & - & 94.9 \\
        
        BR & 95.0 & 95.1 & 94.5 & 91.7 & 94.9 & - & 94.2 \\
        
        \rowcolor{bg_blue}
        LongReMix & 96.3 & 95.1 & 93.8 & 79.9 & 94.7 & 84.4 & 91.9 \\
        
        DMLP & 94.7  & 94.2 & 93.2 & 92.8 & 93.9 & - & 93.7 \\
        
        \rowcolor{bg_blue}
        PSSCL & 96.4 & 95.6 & 93.73 & 92.9 & 93.9 & 86.9 & 94.5 \\
        
        \rowcolor{bg1_blue}
        \textbf{Ours} & 
        $\textcolor{rose_red}{\textbf{97.24}}^{{\bm{\pm}{0.13}}}$ & 
        $\textcolor{rose_red}{\textbf{96.29}}^{{\bm{\pm}{0.10}}}$ & 
        $\textcolor{rose_red}{\textbf{95.94}}^{{\bm{\pm}{0.12}}}$ & 
        $\textcolor{rose_red}{\textbf{93.7}}^{{\bm{\pm}{0.16}}}$ & 
        $\textcolor{rose_red}{\textbf{94.41}}^{{\bm{\pm}{0.09}}}$ & 
        $\textcolor{rose_red}{\textbf{87.91}}^{{\bm{\pm}{0.11}}}$ & 
        $\textcolor{rose_red}{\textbf{95.4}}^{{\bm{\pm}{0.12}}}$ \\ 
        \bottomrule
    \end{tabularx}
\end{table*}

\begin{table}[t]
    \centering
    \caption{Experimental results on CIFAR-100 dataset.}
    \label{tab:cifar100}
    
    \begin{tabularx}{\columnwidth}{l *{3}{Y}}
        \toprule
        Dataset & \multicolumn{3}{c}{CIFAR-100} \\ 
        Noise Mode & \multicolumn{2}{c}{Sym.} & {Asym.}  \\ 
        Method & 20\% & 50\% & 40\%  \\ 
       \midrule
        \rowcolor{bg_gray}
        Standard CE & 62.0 & 46.7 & 44.5  \\
        \rowcolor{bg_gray}
        GCE & 59.2 & 47.8 & 42.9  \\
        \rowcolor{bg_gray}
        GCE+CL & 68.1 & 53.3 & 44.1  \\
        \rowcolor{bg_blue}
        DivideMix & 77.3 & 75.6 & 59.1  \\
        ELR+ & 77.6 & 73.6 & -   \\
        MOIT+ & 75.9 & 70.6 & -  \\
        
        Sel-CL+ & 76.5 & 72.4 & -  \\
        
        \rowcolor{bg_gray}
        MixUp & 67.8 & 57.3 & 48.1  \\
        
        UNICON & 78.9 & 77.6 & -  \\
        
        SOP+ & 78.8 & 75.9 & -  \\
        BR & 69.0 & 68.9 & -  \\
         \rowcolor{bg_blue}
        LongReMix & 77.9 & 75.5 & 59.8  \\
        DMLP & 72.7 & 68.0 & - \\
        
        \rowcolor{bg_blue}
        PSSCL & 77.6 & 77.0 & 61.8 \\
        
        \rowcolor{bg1_blue}
        \textbf{Ours} & 
        $\textcolor{rose_red}{\textbf{79.43}}^{{\bm{\pm}{0.06}}}$ & 
        $\textcolor{rose_red}{\textbf{79.19}}^{{\bm{\pm}{0.09}}}$ & 
        $\textcolor{rose_red}{\textbf{63.53}}^{{\bm{\pm}{0.13}}}$ \\
        \bottomrule
    \end{tabularx}%
    
\end{table}

\section{Experiments}
\subsection{Experimental Setup}
In this section, we benchmark our approach against a comprehensive set of representative methods, including standard CE\cite{CE}, DivideMix\cite{li2020dividemix}, UNICON\cite{unicon}, LongReMix\cite{longremix}, ScanMix\cite{scanmix}, among others\cite{mixup,sop,promix}. We evaluate these methods on two synthetic noise datasets (i.e., CIFAR-10/100\cite{cifar}), which contain both symmetric and asymmetric label noise. Additionally, we utilize multiple real-world datasets. To ensure a fair comparison with existing sample selection methods, almost all experimental settings in this study remain consistent with these methods across each dataset, particularly regarding the backbone architecture, optimizer, and $\lambda_{cl}$. Unless otherwise stated, the value of $q$ in the GCE loss is set to 0.7, $\lambda_{spade}$ is set to 0.1.

\textbf{CIFAR-10/100:} These datasets contain 50K training images and 10K test images. We adopt standard preprocessing strategies. In our experiments, we utilize two types of synthetic noise models: Symmetric noise (Sym), flips labels to other classes with a ratio $r*$, and Asymmetric noise (Asym), where CIFAR-10 follows similar-class mapping, and CIFAR-100 involves circular flipping within superclasses.

\textbf{Animal-10N:} This dataset contains 5 pairs of confusing animals (e.g., cat vs. lynx)\cite{selfie}, with an estimated noise rate of approximately 8\%. As a recognized benchmark dataset in previous LNL research, we select VGG-19N as the backbone network for this dataset.

\textbf{Food-101:} A real-world dataset collected from the web\cite{food}. Consistent with prior studies, we use a ResNet-50 network pre-trained on ImageNet as the backbone.

\subsection{Experimental Results on Synthetic Datasets}
We demonstrate the performance of the proposed paradigm under various label noise scenarios. First, using synthetic noisy label datasets, we consider Sym rates of 20\%, 50\%, 80\%, and 90\%, as well as Asym rates of 40\% and 49\%.

\textbf{CIFAR-10:} Table~\ref{tab:cifar10_results} demonstrates that our method consistently outperforms baseline methods across all noise settings. In the extreme 90\% Sym scenario, we achieve an accuracy of \textbf{93.7\%}, surpassing existing SOTA benchmarks. For the highly challenging 49\% Asym setting where current SOTA methods typically struggle due to severe inter-class confusion, our method maintains a remarkable accuracy of \textbf{87.91\%}. This outperforms LongReMix by a substantial margin of \textbf{3.5\%} and exceeds PSSCL by approximately 1\%.

\textbf{CIFAR-100:} Table~\ref{tab:cifar100} reports the best test accuracy on the CIFAR-100 dataset. Regarding Sym, our method achieves accuracies of \textbf{79.43\%} and \textbf{79.19\%} at noise rates of 20\% and 50\%, respectively. Notably, as the noise ratio escalates to 50\%, while most competing methods suffer noticeable performance degradation, our method demonstrates exceptional stability, with a performance degradation of less than \textbf{0.3\%}. In the more challenging 40\% Asym setting, our approach secures a top accuracy of \textbf{63.53\%}. Compared to the current SOTA method PSSCL, our approach achieves a notable improvement of \textbf{1.73\%}.


\subsection{Experimental Results on Real-World Datasets}

\textbf{Animal-10N:} Table~\ref{tab:animal10n} presents the comparative performance on this real-world noisy label dataset. Our method achieves a SOTA test accuracy of \textbf{88.1\%}. Compared to classical baselines, ours delivers a significant performance boost of over \textbf{6\%}. Crucially, even when pitted against the most advanced competitors such as LongReMix and DICS, our method maintains a robust lead of approximately 1\%. This validates the superiority of actively reshaping the feature space in handling real-world label noise.

\begin{table}[t]
    \centering
    \caption{Experimental results on Animal-10N dataset using VGG-19N."\dag" means ResNet-18 is employed.}
    \label{tab:animal10n}
    \begin{tabularx}{\columnwidth}{l *{2}{Y}}
        \toprule
    
        Method & Ref. & Test Accuracy  \\ 
       \midrule
       \rowcolor{bg_gray}
        Standard CE & \textcolor{cite_blue}{Baseline} & 79.4 \\
        \rowcolor{bg_gray}
        GCE & \textcolor{cite_blue}{NeurIPS18} & 81.5 \\
        Co-teaching+Nested & \textcolor{cite_blue}{NeurIPS18} & 84.1 \\
        MixUp & \textcolor{cite_blue}{ICLR18} & 82.7 \\
        \rowcolor{bg_gray}
        SELFIE & \textcolor{cite_blue}{ICML19} & 81.8 \\
        
        DivideMix & \textcolor{cite_blue}{ICLR20} & 84.5 \\
        
        PLC & \textcolor{cite_blue}{ICLR21} & 83.4 \\
        \rowcolor{bg_gray}
        DAL \dag & \textcolor{cite_blue}{PR23} & 82.7 \\
        BR & \textcolor{cite_blue}{WACV23} & 85.8 \\
        OT-Filter & \textcolor{cite_blue}{CVPR23} & 85.5 \\
        \rowcolor{bg_blue}
        DICS & \textcolor{cite_blue}{CVPR23} & 87.1 \\
        \rowcolor{bg_blue}
        LongReMix & \textcolor{cite_blue}{PR23} & 87.2 \\
        \rowcolor{bg_blue}
        PSSCL & \textcolor{cite_blue}{PR25} & 86.2 \\
        \rowcolor{bg1_blue}
        \textbf{Ours} & \textcolor{cite_blue}{\textbf{-}} &$\textcolor{rose_red}{\textbf{88.1}}^{{\bm{\pm}{0.10}}}$ \\

        \bottomrule
    \end{tabularx}%
    
\end{table}

\textbf{Food-101:} Table~\ref{tab:food101} presents the results on the fine-grained classification task. While current SOTA methods plateau between 86.2\% and 86.4\%, our method achieves an accuracy of \textbf{87.4\%}. This result indicates that when dealing with fine-grained visual confusion and real-world label noise, the proposed virtual outlier synthesis strategy establishes an energy potential around the feature manifold. This effectively reshapes the feature space, thereby significantly reducing feature overlap and confusion among fine-grained classes.

\begin{table}[t]
    \centering
    \caption{Experimental results on Food-101 dataset using pre-trained ResNet-50."\dag" means that ResNet-18 is employed.}
    \label{tab:food101}
    \begin{tabularx}{\columnwidth}{l *{2}{Y}}
        \toprule
    
        Method & Ref. & Test Accuracy  \\ 
       \midrule
       \rowcolor{bg_gray}
        Standard CE & \textcolor{cite_blue}{Baseline} & 81.4 \\
        DivideMix & \textcolor{cite_blue}{ICLR20} & 85.9 \\
        PLC & \textcolor{cite_blue}{ICLR21} & 85.3 \\
        WarPI & \textcolor{cite_blue}{PR22} & 85.9 \\
        NoiseRank & \textcolor{cite_blue}{CVPR23} & 85.2 \\
        SMP & \textcolor{cite_blue}{CVPR23} & 85.1 \\
        \rowcolor{bg_gray}
        CleanNet & \textcolor{cite_blue}{CVPR23} & 83.5 \\

        \rowcolor{bg_gray}
        SPRL\dag & \textcolor{cite_blue}{PR23} & 76.1 \\
        SNSCL & \textcolor{cite_blue}{CVPR23} & 85.4 \\
        \rowcolor{bg_blue}
        SNSCL+ & \textcolor{cite_blue}{CVPR23} & 86.4 \\
        \rowcolor{bg_blue}
        LongReMix & \textcolor{cite_blue}{PR23} & 86.2 \\
        \rowcolor{bg_blue}
        PSSCL & \textcolor{cite_blue}{PR25} & 86.4 \\
        \rowcolor{bg1_blue}
        \textbf{Ours} & \textcolor{cite_blue}{\textbf{-}} & $\textcolor{rose_red}{\textbf{87.4}}^{{\bm{\pm}{0.14}}}$ \\
        \bottomrule
    \end{tabularx}%
    
\end{table}

\begin{table}[htbp]
  \centering
  \caption{Performance comparison of various sampling strategies on CIFAR-10.}
  \label{tab:sampling_strategies}
  \begin{tabularx}{\linewidth}{lYYY}
    \toprule
    Dataset & \multicolumn{3}{c}{CIFAR-10} \\
    
    Noise Mode & \multicolumn{2}{c}{Sym.} & Asym. \\
    
    Sampling Strategy & 20\%  & 80\% & 40\% \\
    \midrule
    Hybrid       & 81.79     & 74.84    & 90.32    \\
    Gaussian     & 82.66     & 77.37    & 91.10    \\
    \rowcolor{bg_blue}
    Perturbation & 94.50     & 91.78    & 90.32    \\
    \rowcolor{bg1_blue}
    \textbf{Uniform (Ours)} & \textcolor{rose_red}{\textbf{97.24}} & \textcolor{rose_red}{\textbf{95.94}} & \textcolor{rose_red}{\textbf{94.41}} \\
    \bottomrule
  \end{tabularx}
\end{table}

\begin{figure*}[t]
    \centering
    \includegraphics[width=1\textwidth]{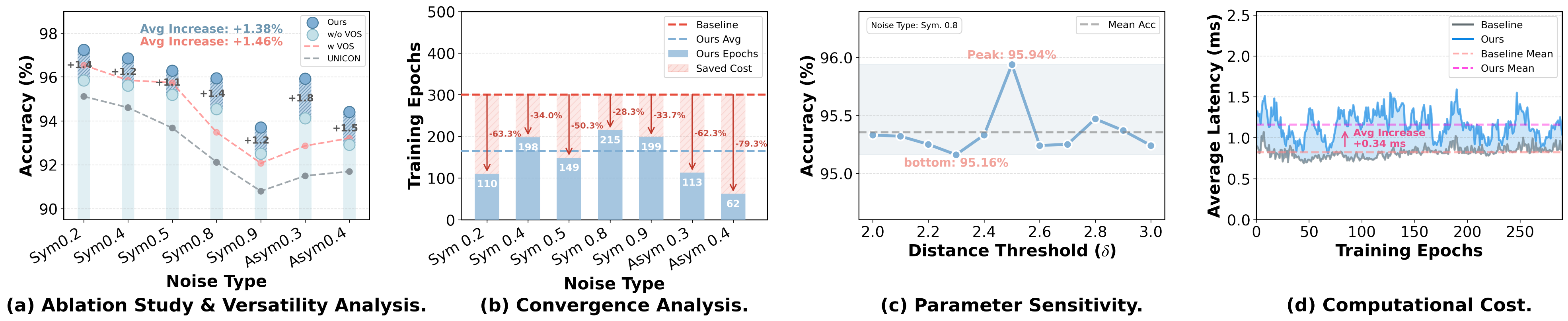} 
    \caption{Detailed analysis of the proposed framework. 
     \textbf{(a):} The bar charts illustrate the contribution of the VOS module to classification accuracy across various noise settings. Meanwhile, the line plots validate its versatility, integrating VOS into the UNICON baseline yields significant performance gains.
     \textbf{(b):} Comparison of training efficiency, demonstrating a significant reduction in the number of epochs required for convergence.
     \textbf{(c):} Robustness analysis of the distance threshold $\tau$, showing stable performance without extensive fine-tuning.
     \textbf{(d):} Benchmarking of average training latency per iteration to assess the overhead introduced by outlier synthesis.}
\label{fig:analysis}
    \label{fig:2}
\end{figure*}

\subsection{Detailed Analysis}
\subsubsection{Ablation on Virtual Synthesis}
We conducted an ablation study to verify the contribution of VOS to classification accuracy . As shown in Figure~\ref{fig:2}(a), experimental results demonstrate that removing VOS leads to an accuracy decrease of \textbf{1.06\%}--\textbf{1.88\%} across all noise settings. This validates the effectiveness of reshaping the feature space via synthesizing virtual outliers.

\subsubsection{Convergence Analysis at Performance Parity with PASSCL}

As illustrated in Figure~\ref{fig:2}(b), compared to the baseline, the incorporation of VOS achieves a substantial reduction in training costs across various noise settings. Under Sym 20\% and 50\% settings, the required training epochs are reduced by \textbf{63.3\%} and \textbf{50.3\%}. Notably, under the challenging Asym 40\% setting, the training overhead plummets by \textbf{79.3\%}. These results compellingly demonstrate that by actively constructing geometric boundaries, the model can precisely distinguish noise without undergoing prolonged fluctuation periods, thereby drastically improving training efficiency while maintaining high performance.

\subsubsection{Parameter Sensitivity}
We investigated the sensitivity of the virtual outlier synthesis to the distance threshold $\tau$. As shown in Figure ~\ref{fig:2}(c), the model demonstrates remarkable stability, with accuracy fluctuating slightly between 95.16\% and 95.94\% and averaging 95.35\%. It is worth noting that the optimal performance of 95.94\% is achieved at $\tau=2.5$. This confirms that our new paradigm is robust against hyperparameter changes and yields superior results without extensive fine-tuning.

\subsubsection{Computational Cost Analysis}
To quantify the computational overhead of the proposed paradigm, a comprehensive training efficiency evaluation is conducted. As illustrated in Figure~\ref{fig:2}(d), while the SOTA method PSSCL maintains an average latency of approximately 0.82 ms per iteration, the incorporation of the VOS module increases the latency to roughly 1.18 ms. This 44\% increment is primarily attributed to the additional computational load from outlier synthesis, sampling, and energy-based scoring. The latency curve for VOS exhibits stochastic fluctuations ranging from 0.9 ms to 1.6 ms, reflecting the dynamic number of effective outliers produced by the distance-threshold-based sampling in each epoch.

Table~\ref{tab:cifar10_training_time_v2} benchmarks the cumulative training time required to reach specific performance milestones. Despite the increased per-iteration cost, the proposed paradigm significantly accelerates global convergence by enhancing feature space separability. Specifically, the average time required for GAMR to achieve the accuracy level of the SOTA method PSSCL is 7.64 hours, representing a 36.8\% reduction compared to PSSCL's 12.09 hours. To reach peak performance, the average duration is 14.90 hours, entailing a reasonable 23.2\% overhead. These results demonstrate that the proposed approach not only outperforms existing SOTA baselines in significantly less time but also establishes a superior performance ceiling with acceptable computational costs.

\begin{table}[htbp]
  \centering
  \caption{Training Time (Hours) Comparison on CIFAR-10.}
  \label{tab:cifar10_training_time_v2}
  \small 
  \setlength{\tabcolsep}{3pt}
  \begin{tabularx}{\linewidth}{l Y Y Y Y Y Y}
    \toprule
    Noise Mode & Sym0.2 & Sym0.5 & Sym0.8 & Sym0.9 & Asym0.4 & Avg. \\
    \midrule
    PSSCL & 11.88 & 11.96 & 12.41 & 12.65 & 11.57 & 12.09 \\
    Ours (to PSSCL) & 7.13 & 7.35 & 10.82 & 9.86 & 3.05 & 7.64 \\
    Ours (Peak) & 14.97 & 14.81 & 15.10 & 14.86 & 14.78 & 14.90 \\
    \bottomrule
  \end{tabularx}
\end{table}

\subsubsection{Versatility Analysis}
It is worth noting that our method does not rely on explicit noise modeling. Instead, by decoupling feature-level optimization from specific filtering logic, it functions as a universal plug-and-play module that can be seamlessly integrated with various sample selection frameworks. We validated this universality by equipping the robust baseline UNICON with our VOS module. As visualized by the curves in Figure~\ref{fig:2}(a), this combination achieves a consistent performance boost, increasing the average accuracy by 2.35\% across all noise scenarios.

\subsubsection{Sampling Strategy Analysis}
We evaluate the efficacy of Uniform sampling against three alternative candidate generation strategies: Perturbation, GMM, and Interpolation-based (Mixup) sampling. Empirical results demonstrate that Uniform sampling exhibits significant superiority, a performance gain that can be attributed to its mitigation of parametric bias sensitivity and its broader exploratory coverage. Detailed analysis is provided in the Appendix.
\begin{figure}[htbp]
    \centering
    \begin{subfigure}[b]{0.22\textwidth}
        \centering
        \includegraphics[width=\linewidth]{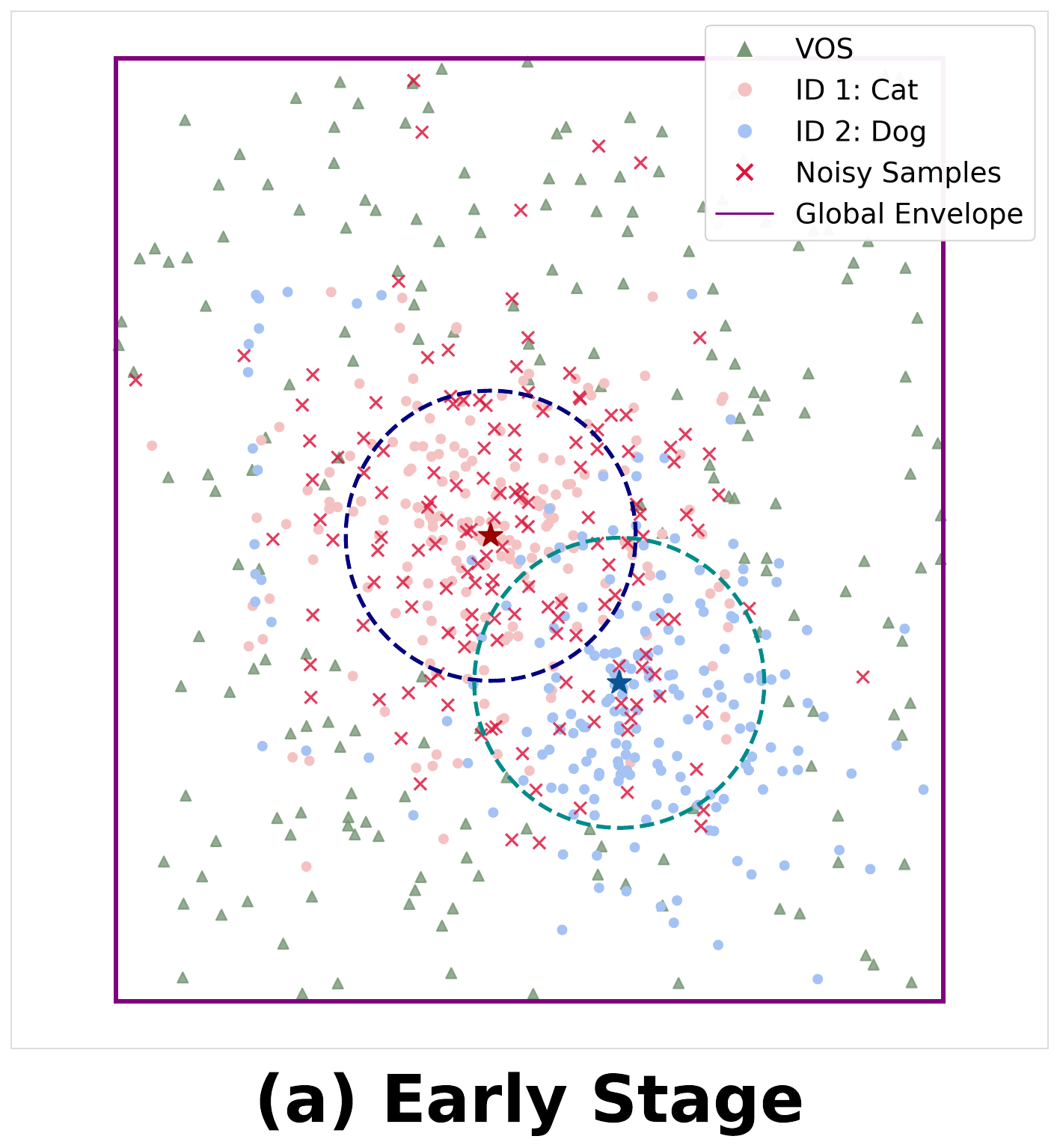} 
        \label{fig:sub1}
    \end{subfigure}
    \hspace{0.05cm} 
    \begin{subfigure}[b]{0.22\textwidth}
        \centering
        \includegraphics[width=\linewidth]{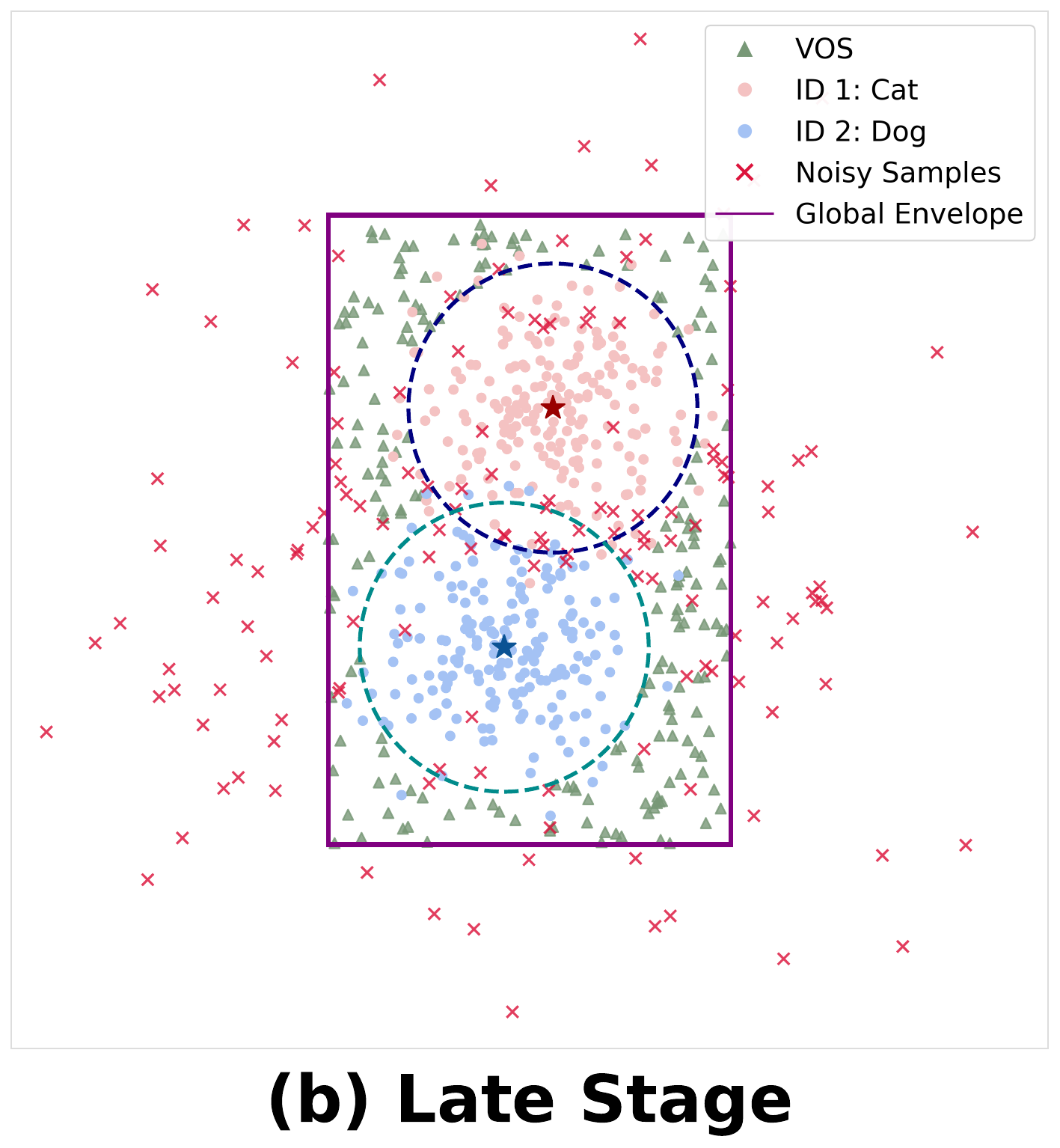} 
        \label{fig:sub2}
    \end{subfigure}
    
    \caption{Manifold Dynamics via Hyper-rectangle Sampling. (a) Early Stage: Loose boundaries offer coarse regularization, stabilizing the feature space against noise. (b) Late Stage: Noise segregation and increased cluster compactness induce the sampling envelope's contraction toward the intrinsic manifold.}
    \label{fig:main}
\end{figure}
\subsection{Visualization of Manifold Dynamics}
To empirically characterize the efficacy of the non-parametric hyper-rectangle sampling paradigm, we visualize the dynamic geometric evolution of feature manifolds and their associated sampling envelopes in Figure ~\ref{fig:main}. As illustrated in Figure ~\ref{fig:main}(a), during the nascent stage of training, ID clusters exhibit a diffused distribution under significant stochastic noise perturbation. In contrast to covariance-based statistics that are prone to outlier-induced distortion, the extreme-value boundaries demonstrate superior geometric robustness, providing fundamental topological constraints for the chaotic feature space. As training progresses, the synthesized VOS samples exert inward centripetal pressure via the energy-based loss, compelling the authentic ID clusters to contract toward their centroids, while mislabeled noise is segregated from the high-density manifold cores into low-density extrinsic regions. The adaptive convergence of these sampling boundaries ultimately fosters a self-purifying cycle, where feature compactness and sampling precision are mutually reinforced.

\begin{table}[t]
    \centering
    \setlength{\tabcolsep}{3pt}
    \caption{Performance comparison of OOD detection on CIFAR-10.}
    \label{tab:cifar10_custom}
    
    \begin{tabularx}{\columnwidth}{l *{4}{Y}}
        \toprule
        Dataset & \multicolumn{4}{c}{CIFAR-10} \\ 
        Noise Mode & Sym0.2 & Sym0.5 & Sym0.8 & Asym0.4  \\ 
        Method &  \multicolumn{4}{c}{AUROC $\uparrow$/FPR95 $\downarrow$}   \\ 
        \midrule
        
      
        PSSCL & 89.31/30.54 & 90.61/37.10 & 87.36/44.32 & 88.00/42.12 \\
        
        UNICON & 89.93/28.77 & 87.78/29.04 & 88.37/35.11 & 88.02/33.44 \\
        
        \rowcolor{bg1_blue}
        \textbf{Ours} & 
        \textcolor{rose_red}{\textbf{93.35/24.04}} & 
        \textcolor{rose_red}{\textbf{92.29/26.74}} & 
        \textcolor{rose_red}{\textbf{90.52/35.30}} & 
        \textcolor{rose_red}{\textbf{91.10/31.83}} \\
        \bottomrule
    \end{tabularx}
\end{table}

\subsection{Noise Robustness in OOD Detection}
Following the strict protocols of the OpenOOD benchmark, we comprehensively evaluated the performance of our proposed paradigm for the OOD detection task against the SOTA baseline PSSCL and the classic baseline UNICON. The evaluation was conducted on five standard OOD datasets, SVHN, LSUN, Places365, Textures, and MNIST, under various noise types and intensities. To quantify the detection performance on unseen data, we adopted two standard metrics: AUROC and FPR95. The results consistently show that across all noise settings, the average AUROC improved by 3.14\%, and the average FPR95 decreased by 9.05\%. This quantitative result directly demonstrates that our method can establish a clearer boundary between ID and OOD data, even when facing severe label noise, thus significantly outperforming existing baseline models.

\section{Conclusion}
In this study, we unveil and address a critical geometric bottleneck often overlooked in SOTA noisy label learning methods. We pioneer a paradigm shift, transitioning the research focus from passive filtering to active reshaping, and instantiate this vision via a universal, energy-based geometric regularization framework. By actively reshaping the feature space to compel the disentanglement of hard samples from noise, our approach enables the model to capture the intrinsic diversity of the data distribution. This provides a fundamental solution to the inherent paradox of performing sample selection within a degenerate feature space, offering a novel geometric perspective for future research.


\bibliographystyle{ACM-Reference-Format}
\bibliography{sample-base}

\end{document}